\documentclass{llncs}
\usepackage{amsbsy}
\usepackage{IEEEtrantools}
\usepackage{multirow}
\usepackage[export]{adjustbox}

\usepackage{floatrow}

\usepackage{amsmath, amssymb, harpoon}
\usepackage{graphicx}
\usepackage[tight,normalsize,sf,SF]{subfigure}
\graphicspath{./}
\usepackage{epstopdf}

\usepackage{fancyhdr}
\DeclareMathOperator*{\dirac}{\delta}
\DeclareMathOperator*{\mydigamma}{\Psi}

\title{Interpretable Low-Rank Document Representations with Label-Dependent\\Sparsity Patterns}

\author{Ivan Ivek}

\institute{Rudjer Boskovic Institute, Bijenicka 54, 10000 Zagreb, Croatia}

\begin{document}
\maketitle
\thispagestyle{plain} 

\begin{abstract}
In context of document classification, where in a corpus of documents their label tags are readily known, an opportunity lies in utilizing label information to learn document representation spaces with better discriminative properties. To this end, in this paper application of a Variational Bayesian Supervised Nonnegative Matrix Factorization (supervised vbNMF) with label-driven sparsity structure of coefficients is proposed for learning of discriminative nonsubtractive latent semantic components occuring in TF-IDF document representations. Constraints are such that the components pursued are made to be frequently occuring in a small set of labels only, making it possible to yield document representations with distinctive label-specific sparse activation patterns. A simple measure of quality of this kind of sparsity structure, dubbed inter-label sparsity, is introduced and experimentally brought into tight connection with classification performance. Representing a great practical convenience, inter-label sparsity is shown to be easily controlled in supervised vbNMF by a single parameter.
\end{abstract}
\section{Introduction}
As an essential step in machine learning applications which both efficiency and quality of learning depend on, dimensionality reduction has become a well covered subject of research \cite{VanDerMaaten2009} which produced archetipal linear methods with low-rank assumptions such as Principal Component Analysis (PCA) \cite{Jolliffe2002} and Nonnegative Matrix Factorization (NMF) \cite{Lee1999a}, as well as their kernelized and generally non-linear variants, to touch upon some. Originally they have been formulated as entirely unsupervised methods. However, in supervised and semi-supervised learning applications, where labels of learning samples are readily available, it may be appealing to use this information to obtain lower-dimensional representations of data which not only attempt to preserve the original variance in the data, but also promise to deliver representation spaces with better discriminative properties. A well known representative which incorporates this desideratum is Fisher's Linear Discriminant Analysis (FLD) \cite{Martinez2001}.\\
In recent relevant literature there is a pronounced  trend of using probabilistic generative models for this purpose. Probabilistic approaches to learning lie on a well developed mathematical apparatus which offers flexible enough modeling of prior knowledge in form of graphical models, supported by well-known meta-algorithms for estimating model parameters. Of this family of algorithms, along with Probabilistic Latent Semantic Analysis (pLSA) \cite{Hofmann1999}, a common probabilistically formulated baseline algorithm in text mining is Latent Dirichlet Allocation (LDA) \cite{Blei2012a} with its more recent discriminative modifications \cite{Blei2007}\cite{Lacoste-Julien2008}, as well as probabilistic formulations of sparse NMF \cite{Cemgil2009} and their supervised counterparts \cite{Ivek2014}.\\
Sparse coding is known to result in efficient and robust representations which have proven suitable for applications such as data compression, denoising and missing data imputting \cite{Mallat1999}. On the other hand, representations obtained discriminatively are suitable for classification purposes. Combining those two properties, the basis of this work is a probabilistically formulated method for sparse additive representations of data using nonnegative latent components which are of sparsity structure additionally driven by data labeling \cite{Ivek2014}. In context of document classification, the decomposition is suitable for finding interpretable patterns of semantically related terms, with high discriminative potential.
\subsection{Document Feature Spaces}
Disregarding syntactic and semantic interrelations of words, the simplest and most often used intermediate form for document representation is bag-of-words; after tokenization, purification and stemming, frequency of relevant terms is  determined for each document resulting in representations of documents as frequencies of particular terms.
Models such as LDA have a natural interpretation when decomposing bag-of-words representations, while other approaches may benefit from TF-IDF weighting \cite{Rajaraman2011} which heuristically measures the importance of a term for a particular document in a specific corpus of documents. For a term with index $\tau$ in $\nu$-th document, as a product of two measures,
\begin{equation}
	tfidf_{\nu\tau} = tf_{\nu\tau}*idf_\tau,
\end{equation}
TF-IDF score is proportional to (normalized) frequency of a particular term in a document,
\begin{equation}
	tf_{\nu\tau}=\frac{\#_{\nu\tau}}{\max_t(\#_{\nu t})},
\end{equation}
but stunted by a measure of how rare this term occurs in the entire corpus,
\begin{equation}
	idf_\tau=\ln \frac{N}{n_\tau},
\end{equation}
where the number of occurences of term $\tau$ in $\nu$-th document is denoted by $\#_{\nu\tau}$, the number of documents in the corpus by $N$ and the number of documents which contain term $\tau$ at least once by $n_\tau$.
\subsection{NMF as a Tool for Latent Semantic Analysis}
Bag-of-words-based approaches to text mining are known to suffer from problems of polysemy and synonymy of terms. These problems can be alleviated by representing documents in spaces of patterns of frequencies of semantically related terms rather than in the original space of term frequencies \cite{Deerwester1990}. Luckily, algorithms for learning of such representations exist, of which perhaps the best known are pLSA formulations. Also assuming inherent nonnegativity in the data, NMF decompositions can be interpreted the same way as pLSA, revealing patterns of semantically related terms underlying the data. Furthermore, a specific connection worth mentioning is that a NMF formulation based on generalized KL-divergence minimizes the exactly same objective function as the original pLSA formulation does \cite{Goutte2005a}. \\
Nonnegativity is a reasonable assumption and a desireable bias when modeling either term frequencies or derived intermediate document representations such as TF-IDF. In general, NMF aims at decompositions in form of $\boldsymbol{X}{\approx}\mathit{\boldsymbol{TV}}$, where $\boldsymbol{X}$, $\boldsymbol{T}$ and $\boldsymbol{V}$ are all nonnegative matrices. Although the decomposition is nonunique in general, to some extent nonuniqueness may be compensated for by adding additional bias in the model, of which most prominent is sparsity of solution \cite{Lee1999a}. Sparsity is enforced in divergence-based NMF by different sparsity promoting regularizers, e.g. \cite{Hoyer2004}, and in probabilistic formulations by imposing sparse prior distributions on the coefficients \cite{Cemgil2009}.\\
Throughout this paper, in context of document representation for categorization purposes, $\boldsymbol{X}$ will be regarded as a collection of documents organized columnwise and represented by TF-IDF features, $\boldsymbol{T}$ as a low-rank collection of latent semantic components organized columnwise, and $\boldsymbol{V}$ as matrix of coefficients when $\boldsymbol{X}$ is projected onto the space of latent semantic components $\boldsymbol{T}$.  In other words, each document is modeled as a strict superposition of the nonnegative latent semantic components.
\section{Methodology}
\subsection{Supervised NMF Model}
The generative model \cite{Ivek2014} assumes that each column of data, $x_{: \tau}$, is a result of latent components $t_{: i}$ consisting of independent gamma-distributed variables,
\begin{equation}
	p\left(t_{\nu i} \middle| a_{\nu i}^{t},b_{\nu i}^{t}\right)= \mathcal{G}\left(t_{\nu i}\middle|a_{\nu i}^{t},b_{\nu i}^{t}\right),
\end{equation}
interacting through linear mixing with coefficients $v_{i \tau}$ under Poissonian noise:
\begin{equation}
	p\left(s_{\nu i \tau}\middle|t_{\nu i},v_{i \tau}\right)=\mathcal{P}\left(s_{\nu i \tau}\middle|t_{\nu i},v_{i \tau}\right)
\end{equation}
\begin{equation}
	p\left(x_{\nu \tau}\middle|s_{\nu : \tau}\right)=\dirac\left(x_{\nu \tau}-\sum _i{s_{\nu i \tau}} \right).
\end{equation}
Mixing coefficients $v_{i\tau}$ are assumed to be exponentially distributed with different scale parameters for different selections of label indicators $z_\tau \in\mathcal{L}$, formulated as mixtures of variables $\lambda_{i l}$ with $z_\tau$ as discrete numerical mixture selection variables,
\begin{equation}
	p\left(v_{i \tau}\middle|z_\tau,\lambda_{i:}\right)=\mathcal{G}\left(v_{i \tau}\middle|1,\sum_{l\in\mathcal{L}} \dirac(z_\tau-l)\lambda_{i l}^{-1}\right)
\end{equation}
Note that label indicators $z_\tau$ are elements of a discrete set of (integer) numbers $\mathcal{L}$ for convenience of notation. Variables $\lambda_{i l}^{-1}$, representing expectations of magnitudes of coefficient components $i$ for all samples labeled as $l$ are constrained by inverse-gamma priors,
\begin{equation}
	p\left(\lambda_{il}\middle| a_{i l}^{\lambda},b_{i l}^{\lambda} \right)=\mathcal{G}\left(\lambda_{il}\middle| a_{i l}^{\lambda},b_{i l}^{\lambda} \right).
\end{equation}
Because inverse-gamma is a heavy-tailed distribution, by setting the probability mass to be concentrated around some small value, significantly larger values of $\lambda_{i l}^{-1}$ will occur rarely. Thus, such a prior imposes an additional bias to produce models having only a minority of indicators $\lambda_{i l}^{-1}$ with significantly large mean values on average, which, hierarchically propagating to activation coefficients $v_{i\tau}$, constrain samples having the same label to have only a small shared subgroup of latent patterns significantly active.\\
Using compact notation
\begin{IEEEeqnarray}{c}
	p\left(\boldsymbol{X}\middle|\boldsymbol{S}\right)=\prod_{\nu,\tau}p\left(x_{\nu\tau}\middle|s_{\nu:\tau}\right)\nonumber\\
	p\left(\boldsymbol{S}\middle|\boldsymbol{T},\boldsymbol{V}\right)=\prod_{\nu,\tau} p\left(s_{\nu:\tau}\middle|t_{\nu i}, v_{i\tau}\right)\nonumber\\
	p\left(\boldsymbol{T}\middle|\boldsymbol{A^t},\boldsymbol{B^t}\right)=\prod_{\nu,i}p(t_{\nu i}|a_{\nu i}^t,b_{\nu i}^t)\nonumber\\
	p\left(\boldsymbol{V}\middle|\boldsymbol{\Lambda},\overset{\rightharpoonup }{z} \right)=\prod_{i,\tau}p\left(v_{i\tau}\middle|\lambda_{i:},z_\tau\right)\nonumber\\
	p\left(\boldsymbol{\Lambda}\middle|\boldsymbol{A^\lambda},\boldsymbol{B^\lambda} \right)=\prod_{i,l}p\left(\lambda_{il}\middle|a_{il}^\lambda,a_{il}^\lambda \right),\nonumber
\end{IEEEeqnarray}
joint distribution of the supervised NMF model can be written as
\begin{IEEEeqnarray}{rCl}
	\IEEEeqnarraymulticol{3}{l}{
		p\left(\boldsymbol{X},\boldsymbol{S},\boldsymbol{T},\boldsymbol{V},		\boldsymbol{\Lambda}\middle|\boldsymbol{A^t},\boldsymbol{B^t},\boldsymbol{A^\lambda},\boldsymbol{B^\lambda},\overset{\rightharpoonup }{z}\right)
	}\nonumber\\
	\quad &=& p\left(\boldsymbol{X}\middle|\boldsymbol{S}\right)p\left(\boldsymbol{S}\middle|\boldsymbol{T},\boldsymbol{V}\right)p\left(\boldsymbol{T}\middle|\boldsymbol{A^t},\boldsymbol{B^t}\right)
	p\left(\boldsymbol{V}\middle|\boldsymbol{\Lambda},\overset{\rightharpoonup }	{z}\right)p\left(\boldsymbol{\Lambda}\middle|\boldsymbol{A^\lambda},\boldsymbol{B^\lambda}\right).
\end{IEEEeqnarray}
Linear mixing as described by (4), (5) and (6) is the same as in Poisson-gamma NMF \cite{Cemgil2009}. Equations (7) and (8) additionally formulate a sparsity structure abstracted from the level of data samples to the level of labels, making it possible to pursue decompositions with recognizable sparsity patterns characteristic of data samples which share the same label tag.
\subsection{Variational Bayesian Learning Algorithm}
To give a concise outline of general treatment of learning by VB, let the observed variables be denoted by $\boldsymbol{D}$, the hyperparameters of a model by $\boldsymbol{H}$ and both the unobserved variables and the model parameters by $\boldsymbol{\Theta}$. Minimization of discrepancy between posterior $p\left(\boldsymbol{\Theta}\middle|\boldsymbol{D},\boldsymbol{H}\right)$ (which is in general difficult to optimize directly, especially in a fully Bayesian manner) and an introduced instrumental approximation $q(\boldsymbol{\Theta})$ measured by Kullback-Liebler divergence gives rise to a lower bound on the posterior,
\begin{equation}
	\mathcal{L}=\left<\ln p\left(\boldsymbol{D},\boldsymbol{\Theta}\middle|\boldsymbol{H}\right)\right>_{q\left(\boldsymbol{\Theta}\right)}+\mathcal{H}\left[q(\boldsymbol{\boldsymbol{\Theta}})\right],
\end{equation}
where entropy of the probability density function in the argument is denoted by $\mathcal{H}\left[.\right]$. Supposing that $q(\boldsymbol{\Theta})$ is of factorized form $q(\boldsymbol{\Theta})=\prod_{\alpha \in C} q(\boldsymbol{\Theta_\alpha})$, it can be shown that the iterative local updates at iteration (n+1) alternating over C in form of
\begin{equation}
	q(\boldsymbol{\Theta_\alpha)}^{(n+1)}\propto \exp\left(\left\langle\ln p\left(\boldsymbol{D},\boldsymbol{\Theta}\middle|\boldsymbol{H}\right)\right\rangle_{\frac{q(\boldsymbol{\Theta})^{(n)}}{q(\boldsymbol{\Theta_\alpha})^{(n)}}}\right)
\end{equation}
improve the lower bound (10) monotonically. Moreover, should the model be conjugate-exponential, for a fully factorized approximation, expressions in (11) neccessarily assume analytical forms \cite{Winn2003a}.\\
For the model (9) variational Bayesian update expressions are derived from (11) by specifying $p\left(\boldsymbol{D},\boldsymbol{\Theta}\middle|\boldsymbol{H}\right)=p\left(\boldsymbol{X},\boldsymbol{S},\boldsymbol{T},\boldsymbol{V},\boldsymbol{\Lambda}\left|\boldsymbol{A}^{\boldsymbol{t}}\right.,\boldsymbol{B}^{\boldsymbol{t}},\boldsymbol{A}^{\boldsymbol{v}},\boldsymbol{B}^{\boldsymbol{v}}\right)\nonumber$ together with instrumental distribution $q\left(\boldsymbol{\Theta}\right)=q\left(\boldsymbol{S},\boldsymbol{T},\boldsymbol{V},\boldsymbol{\Lambda}\right)$, appropriatedly factorized as $q\left(\boldsymbol{S},\boldsymbol{T},\boldsymbol{V},\boldsymbol{\Lambda}\right)=\prod_{\nu,\tau} q\left(s_{\nu : \tau}\right)\prod_{\nu,i} q\left(t_{\nu i}\right)\prod_{i,\tau} q\left(v_{i\tau}\right)\prod_{i,l}q\left(\lambda_{il}\right)$ for computational convenience.
Still, the lower bound according to (10) includes a difficult term related to optimization of $q\left(\lambda_{il}\right)$. For this reason, the lower bound has been relaxed using Jensen's inequality and optimization is done with respect to this relaxed bound \cite{Ivek2014}. An outline of the treatment of the learning algorithm can be found in Appendix B.
\section{Experiments}
All experiments have been performed on \textit{20Newsgroups}\footnote{Available from Jason Rennie's web Page, http://qwone.com/\char`\~jason/20Newsgroups/} dataset, \textit{bydate} version split into training and test sets; rather than estimating the generalization error of classification by crossvalidation techniques, the underlying ambition is merely to explore peak potentials of classification using different representation spaces evaluated on a single train-test split in same conditions.
\subsection{Dataset}
Experiments have been performed on \textit{20Newsgroups} dataset sorted by date wih duplicates and headers removed, with documents having multiple labels left out and preprocessed to obtain a bag-of-words representation. The dataset is split into a training set and a test set.
To alleviate computational load, the set of features has been heuristically reduced to $10000$ terms, based on maximum TF-IDF score accross all documents.
\subsection{Experimental Setup}
Representation spaces in which consequently classification takes place which are taken under consideration are the ones obtained by PCA, Poisson-gamma unsupervised vbNMF \cite{Cemgil2009}, and the supervised vbNMF, all decomposing the matrix of TF-IDF scores of the training set only. Having learned a specific space of reduced dimensionality, representation of the test set in this space is found by projection on the vector basis in case of PCA or by optimizing the matrix of coefficients only using Poisson-gamma vbNMF formulation (i.e. the matrix of latent components is fixed to what has been learned in the training step) in case of both unsupervised and supervised vbNMF methods.\\
For Poisson-gamma vbNMF sparse decompositions have been pursued by fixing shape hyperparameters of the gamma distributed coefficients to a value less than or equal to $1$ throughout the entire run, while other hyperparameters (constrained to be the same for all elements of matrices $\boldsymbol T$ and $\boldsymbol V$, a single one for each of the matrices) have been optimized automatically by maximization of the lower bound directly in a non-Bayesian manner \cite{Cemgil2009}. For supervised vbNMF, hyperparameters aLambda have been fixed and varied, while other hyperparameters have been left to the algorithm to optimize, by direct optimization as in \cite{Cemgil2009}. Specifically, following initialization, $\lambda$ parameters are chosen to be all equal and fixed for a burn-in period of 10 iterations, not until after which they start to get optimized according to the algorithm in Table 3.\\
To accentuate the predictive potentials of the considered representation spaces by themselves, rather than in conjunction with a strong classifier, the classifier of choice is k-NN using cosine similarity metric, with k chosen heuristically as the square root of the cardinality of the training set.\\
Dimension of space of latent components has been varied as a parameter for all decomposition methods. Because at each run the NMF algorithms converge to some local minimum, to explore these local minima, for each parameter set they have been run 10 times with random initializations.
\subsection{Evaluation}
Metrics of classification performance used in the experiments are micro-averaged accuracy, defined as
\begin{equation}
a^{micro}=\frac{\sum_l N^{correct}_l}{\sum_l N^{all}_l},\nonumber
\end{equation}
and macro-averaged accuracy,
\begin{equation}
	a^{macro}=\frac{1}{L}\sum_l {\frac{N^{correct}_l}{N^{all}_l}},\nonumber
\end{equation}
where the number of correctly classified documents belonging to the $l$-th label is denoted by $N^{correct}_l$, the number of documents belonging to $l$-th label in the test split by $N^{all}_l$ and the number of labels by $L$.
By averaging the accuracies calculated separatedly for each of the labels, macro-averaged accuracy compensates for label-imbalance of test datasets.\\
As a measure of sparsity, Hoyer's measure \cite{Hoyer2004}, based on ratio of l1 and l2 norms and originally introduced in context of NMF penalization, will be used. For a vector $\overset{\rightharpoonup }{x}=\left[x_1,...,x_n\right]^T$ it is defined as
\begin{equation}
	sparsity\left(\overset{\rightharpoonup }{x}\right)=\frac{1}{\sqrt{n}-1} \left(\sqrt{n}-\frac{\sum_i\left|x_i\right|}{\sqrt{(\sum_i x_i^2)}}\right),
\end{equation}
taking value of 1 in case only a single element is non-zero (maximum sparsity), and a value of 0 if all elements are equal (minimum sparsity). For the purpose of this paper, when referring to sparsity of matrices, matrix is assumed to be vectorized first by appending its columns, then treating it as a vector according to (12).\\
If labels in a document corpus are meaningfully assigned based on topics of documents, then meaningful discovered latent semantic components are expected to have specific patterns of occurence for documents belonging to a specific label. Using supervised vbNMF, those patterns are modeled as patterns in sparsity of coefficients (i.e. in patterns of support of sparse coefficients) that documents labeled the same have in common. To measure the consistency of occurence of sparsity patterns in labels, let a representation by coefficients of $N$ documents in $I$ dimensional space be denoted by $\boldsymbol{V}\in \mathbb{R}^{IxN}$, i.e. $n$-th document is represented by coefficient vector $[\boldsymbol{V}]_{:n}$, and let sums of coefficient sets which share the same label be accumulated in matrix $\boldsymbol{L}\in \mathbb{R}^{IxL}$, where $L$ is number of labels as
\begin{equation}
	[\boldsymbol{L}]_{:l}=\sum_{n\in N_l}  [\boldsymbol{V}]_{:n}, 
\end{equation}
where $n$ iterates over subset of document indices with the same label, $N_l$.\\
Now, inter-label sparsity can be introduced, defined as sparsity of matrix $\boldsymbol{L}$. The motivation behind (13) is that $l0$ norm of a sum of vectors with the same sparsity pattern (same support) is the same as the exclusive $l0$ norm of such vectors by themselves, and, the more those vectors deviate from the pattern (i.e. when the vectors have differing supports), the larger the $l0$ norm of the sum will be. Note that the latter rationale holds exactly for $l0$ definition of sparsity, while for more relaxed definitions of sparsity such as (12) the behavior will be only qualitatively similar. For the purpose of this paper, sparsity of (13) will be measured as Hoyer's sparsity (12).
\subsection{Results and Discussion}
For comparison, as a baseline, classification results of PCA are plotted against the dimension of representation space on Fig. 1. For unsupervised vbNMF, micro-averaged accuracies averaged accross random initializations for different shape parameters with varying number of latent semantic components are shown in Fig. 2.
\begin{figure}
\centering
\includegraphics[height=2.5478in]{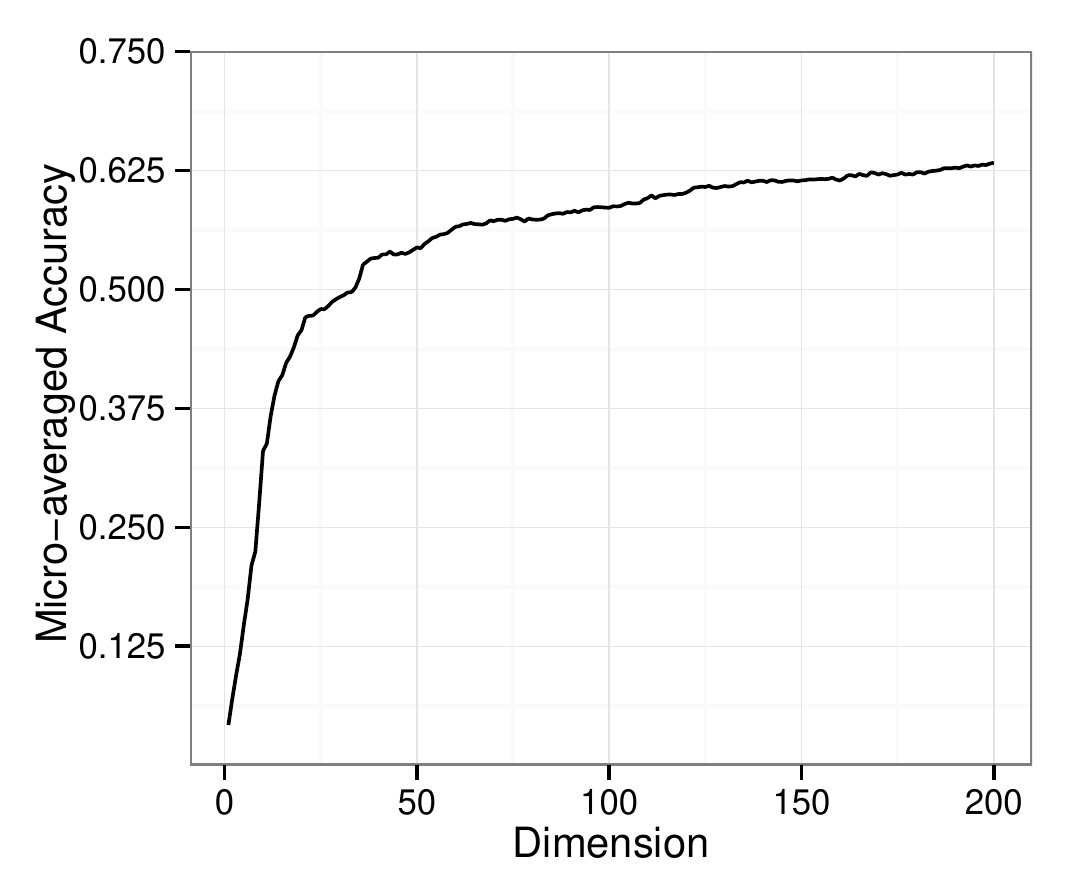}
\caption{Classification results using PCA.}
\label{fig_sim}
\end{figure}
\begin{figure}
\centering
\includegraphics[height=2.5478in]{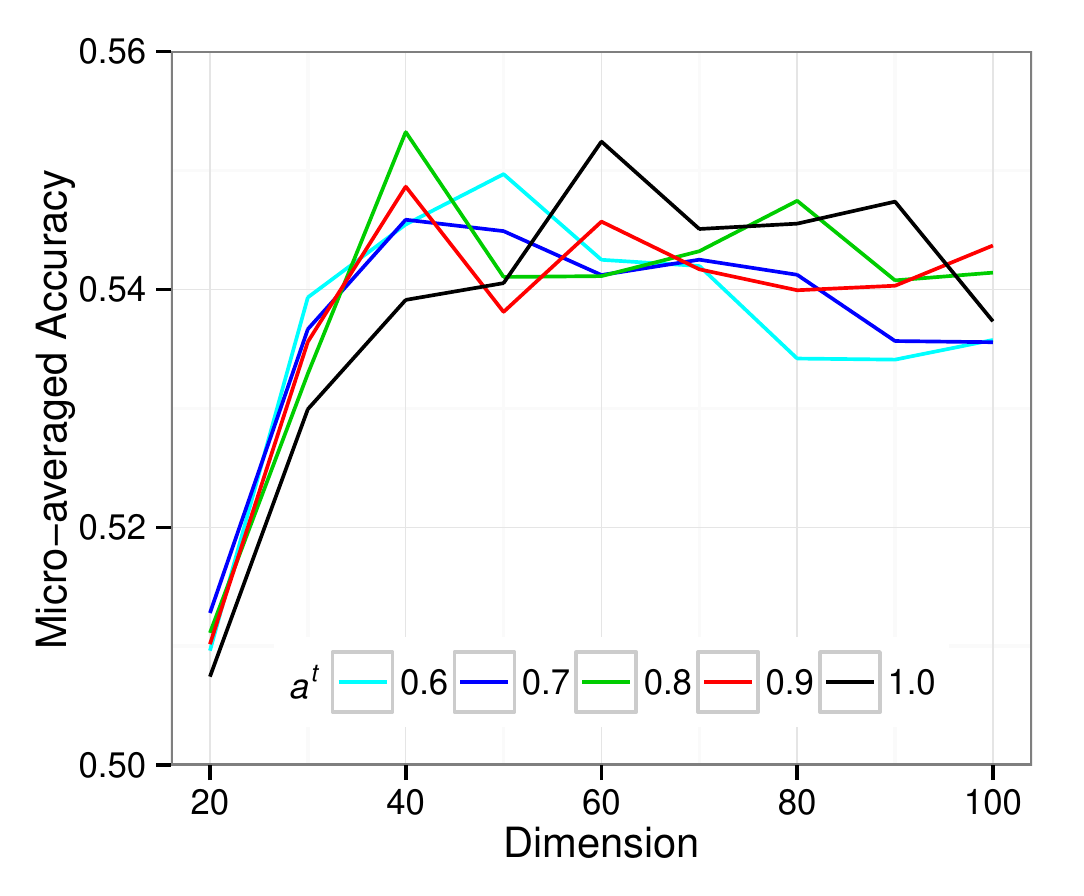}
\caption{Classification results of unsupervised vbNMF (averaged accross 10 random initializations) with varying level of sparsity penalization.}
\label{fig_sim}
\end{figure}
\\
Compared to PCA, even with a larger dimension of representation space, vbNMF with sparsity constraints did not bring improvements on average, regardless of the degree of sparsity penalization. The explanation is that, even though sparse representation spaces may be good for clustering, natural clusters may differ greatly from labeling and consequently even be detrimental to classification applications \cite{Blei2007} when compared to dense representations such as PCA. Better representations for classification purposes are expected to be found by introducing label information to the model, which in spaces obtained by supervised vbNMF (Fig. 3.) indeed manifested as a boost in classification performance.%
\begin{figure*}[]
     \begin{center}
        \subfigure[]{%
            \label{fig:first}
            \includegraphics[height=2.5478in]{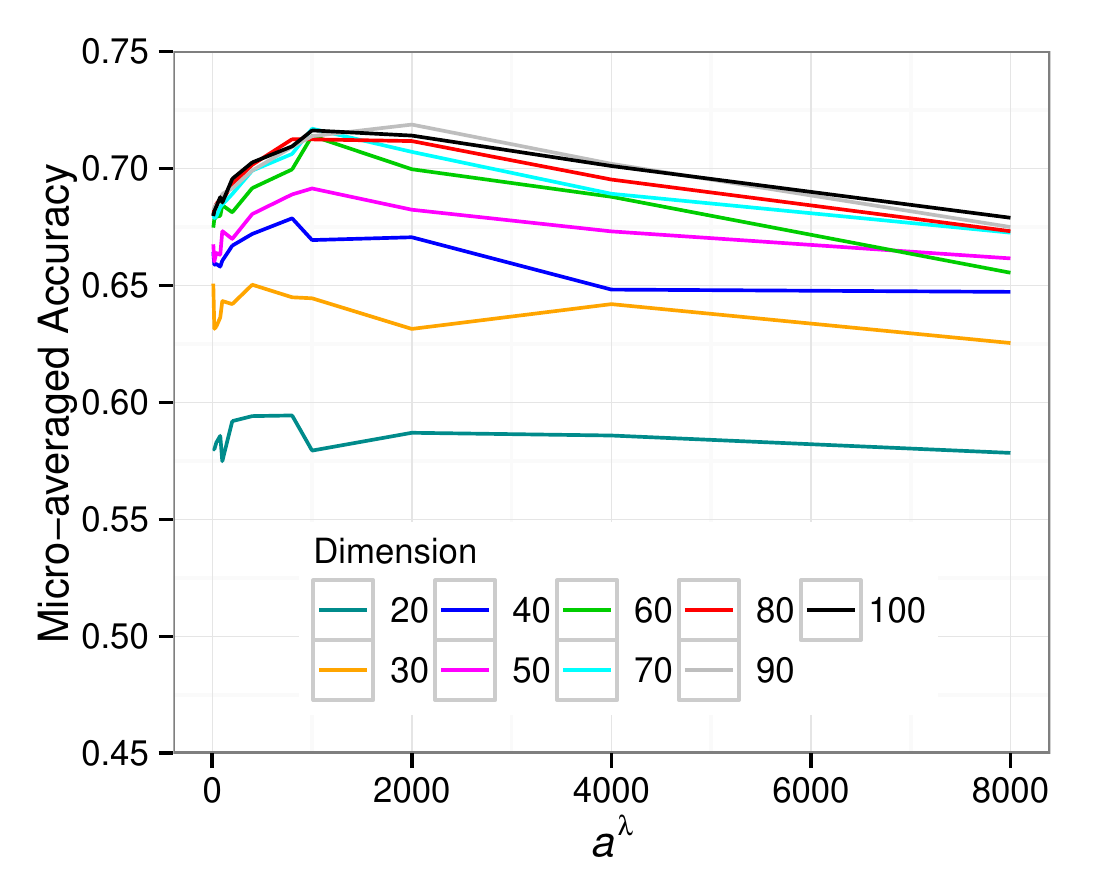}
        }\\
        \subfigure[]{%
           \label{fig:second}
           \includegraphics[height=2.5478in]{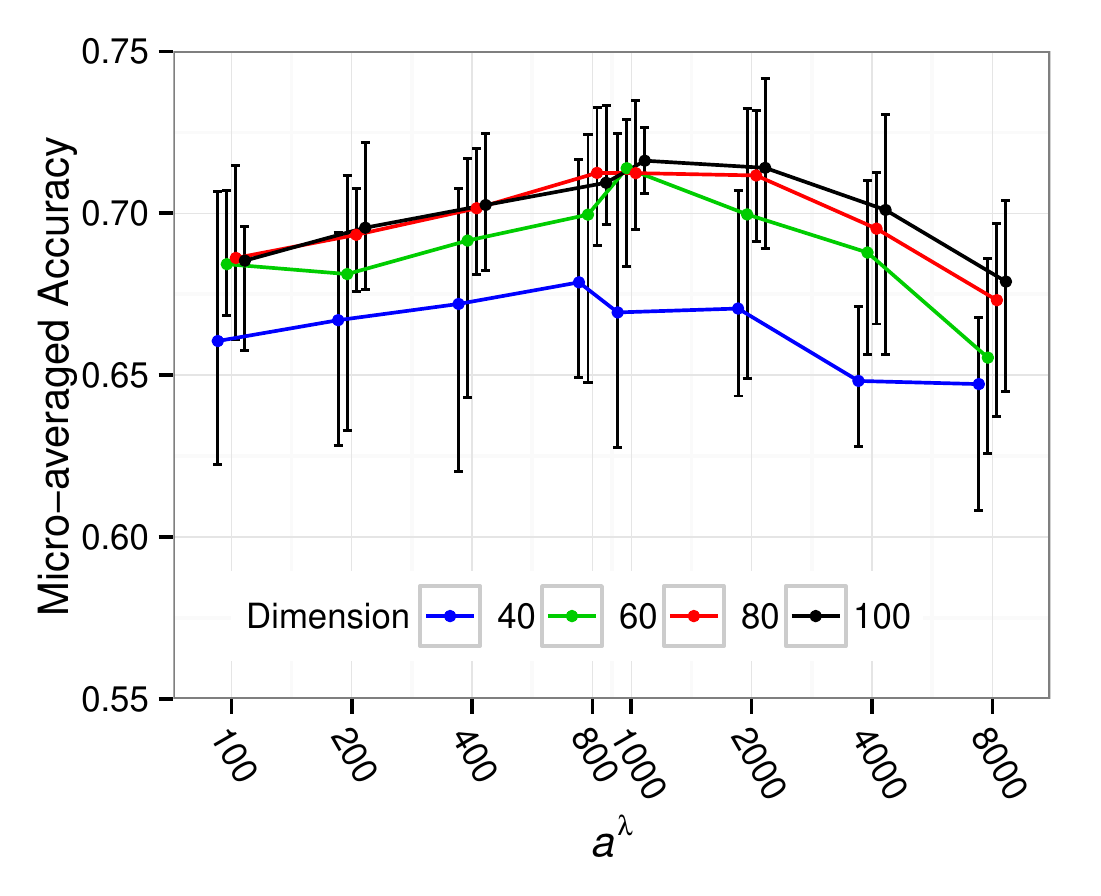}
        }
    \end{center}
    \caption{
Supervised vbNMF classification results, averaged over 10 random initializations. a) Dependence on level of sparsity penalization, varying dimensions of representation spaces. b) Dependence on level of sparsity penalization, varying dimensions of representation spaces. Error bars represent maximum and minimum values among the random initializations. x-axis is shown on logarithmic scale.
     }
   \label{fig:subfigures}
\end{figure*}\\
Both unsupervised vbNMF and supervised vbNMF consistently resulted in sparse decompositions. However, label-driven structure present in supervised vbNMF decompositions (engineered as to be the sole difference in the experiments) is to be accounted for the beneficial effect observed. Examples of sparsities accross labels according to (13) are visualized on on Fig. 4. for the sparse unsupervised vbNMF decomposition which produced peak micro-averaged accuracy of 0.5796 and on Fig. 5. for an arbitrarily chosen supervised variant with matching dimension. The supervised variant produced distinctive sparsity patterns accross labels, which is also reflected quantitatively on inter-label sparsity of the decomposition.\\
The connection between sparsity on the level of labels and classification performance is further explored using Fig. 6., showing data for all supervised representations obtained in the experiments. Variance of the scatter plot becomes tighter with increasing the dimension of representation space, meaning that for a sufficiently large dimension of the decomposition, inter-label sparsity is indeed a good predictor for classification quality on this dataset.\\
Equally importantly, experiments show that in case of supervised vbNMF, inter-label sparsity can elegantly be controlled by a single parameter $a^\lambda$ alone, regardless of dimension: as illustrated by Fig. 7., a logarithmic increase of $a^\lambda$ is accompanied by a trend of growth of inter-label sparsity, only to be broken by too extreme regularizations, when tails of the prior have little mass. On the other hand, unsupervised vbNMF resulted in moderate levels of inter-label sparsity because sparsity structure is supported by the structure of data features only, with somewhat higher values in cases of very strong sparsity regularizations and an impractically small number of latent patterns.\\
Classification results using k-NN classifier with heuristically chosen $k$ on representation spaces obtained by the three methods are summarized in Table 1., reporting peak value of its micro- and macro-averaged accuracies; for the set of parameters which yielded the peak performance, corresponding accuracies averaged accross the 10 random initializations together with minimal achieved accuracies are reported. On Fig. 3.b), showing smooth dependence of micro-averaged accuracy (averaged across random initializations) on an interesting range of $a^\lambda$ for a selection of dimensions, peak performance as entered in Table 1. can be noticed marked.
\begin{figure}
\centering
	\includegraphics[height=5.8in]{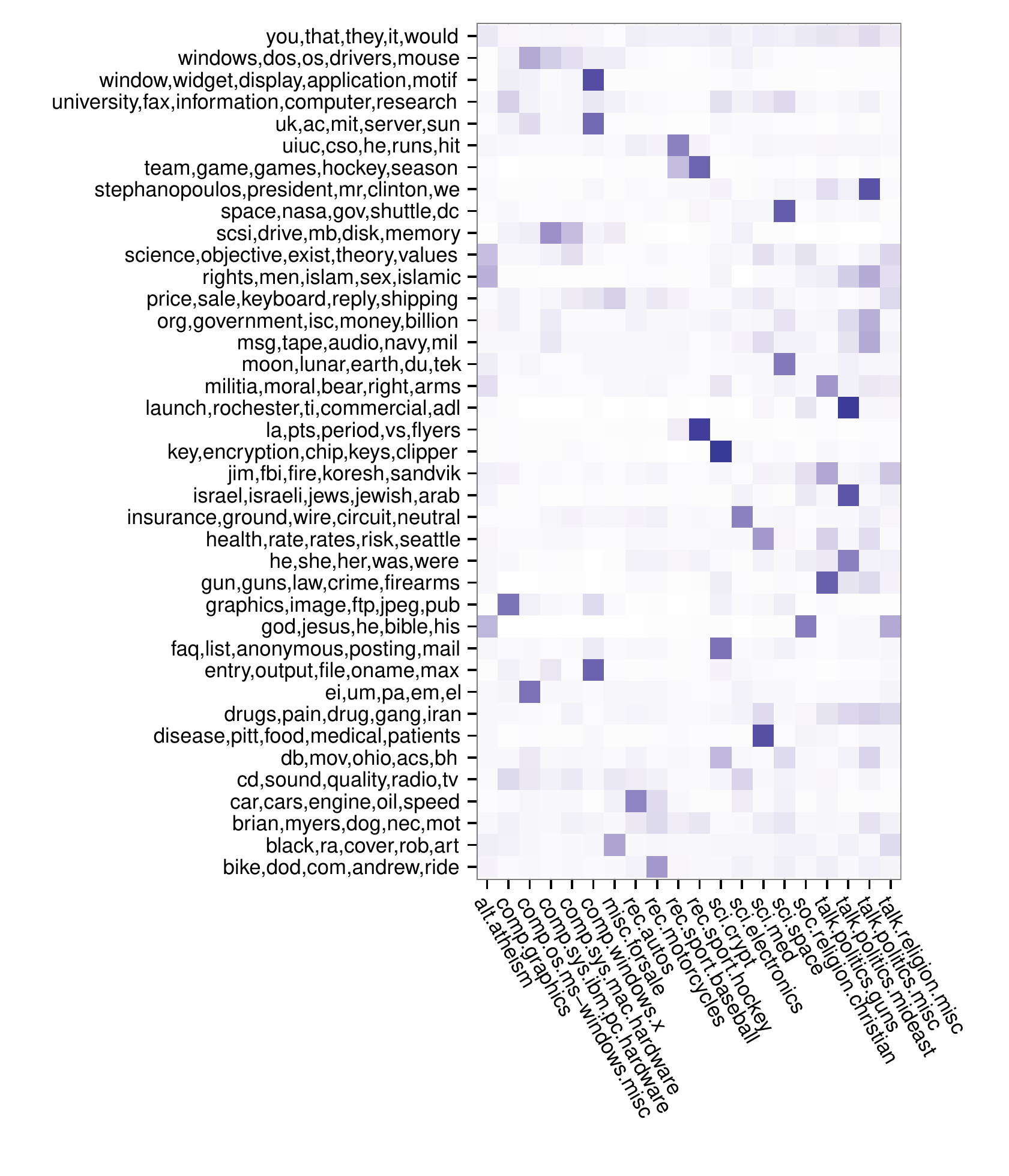}
	\caption{%
Sparsity accross labels according to (13) for the unsupervised vbNMF decomposition with best classification performance; the more intense the tone of blue, the higher the sparsity. Each of the 40 latent semantic components is represented by its 5 most significant terms. Coefficient sparsity is 0.6862, inter-label sparsity is 0.5784. }
\label{fig_sim}
\end{figure}
\begin{figure}
\centering
	\includegraphics[height=5.8in]{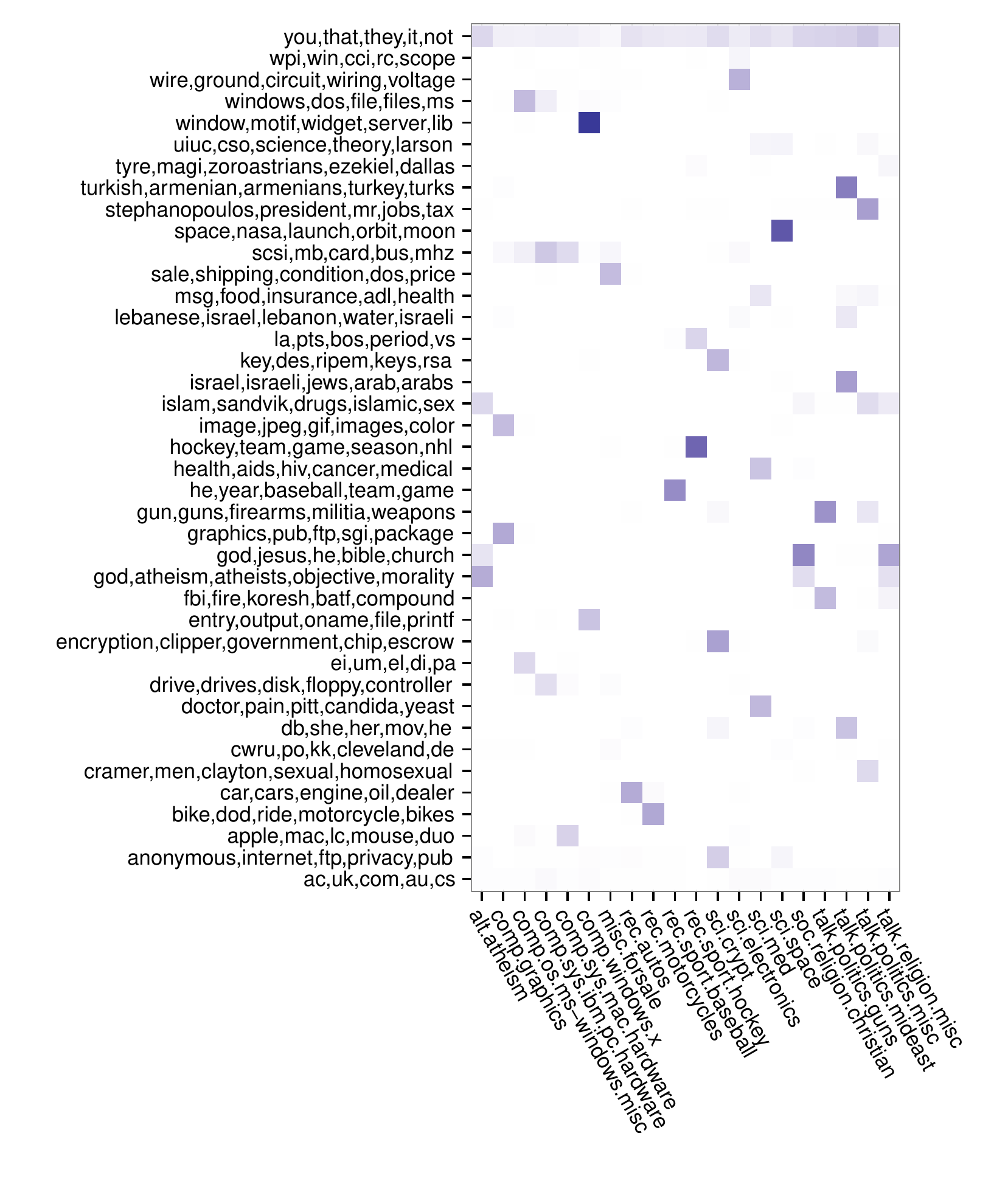}
	\caption{%
		Sparsity accross labels according to (13) for an arbitrarily chosen supervised vbNMF decomposition with 40 latent semantic components, each represented by its 5 most significant terms. Coefficient sparsity is 0.8752, inter-label sparsity is 0.8578.}
\label{fig_sim}
\end{figure}
\begin{figure}
\centering
\includegraphics[height=2.5478in]{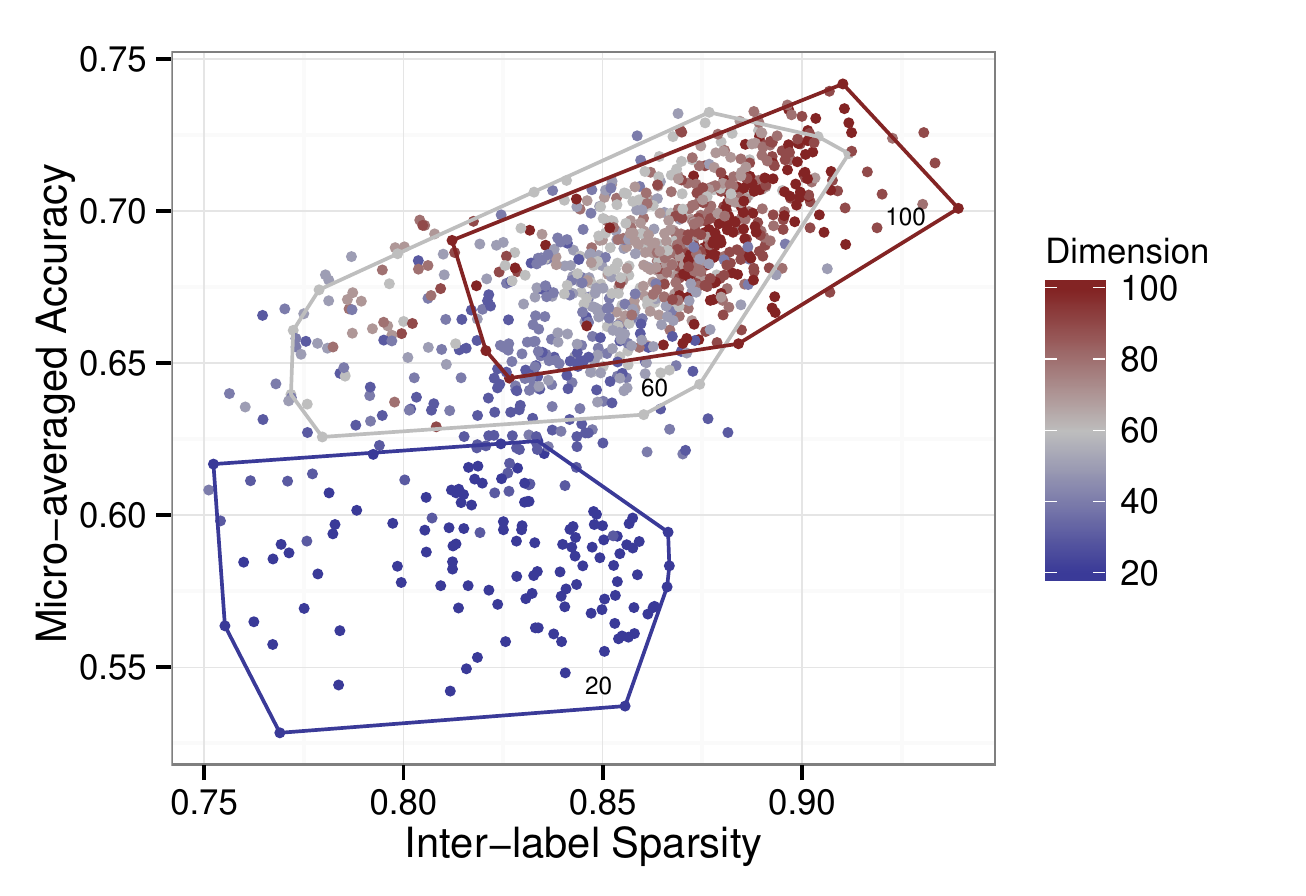}
\caption{Correlation between micro-averaged accuracy and inter-label sparsity. Each point in the scatter plot represents a single supervised NMF decomposition. Convex hulls contain points corresponding to choices of dimensions of 20, 60 and 100.}
\label{fig_sim}
\end{figure}
\begin{figure}[h]
     \begin{center}
        \begin{subfigure}[]{%
            \label{fig:first}
            \includegraphics[height=3.1in,valign=t]{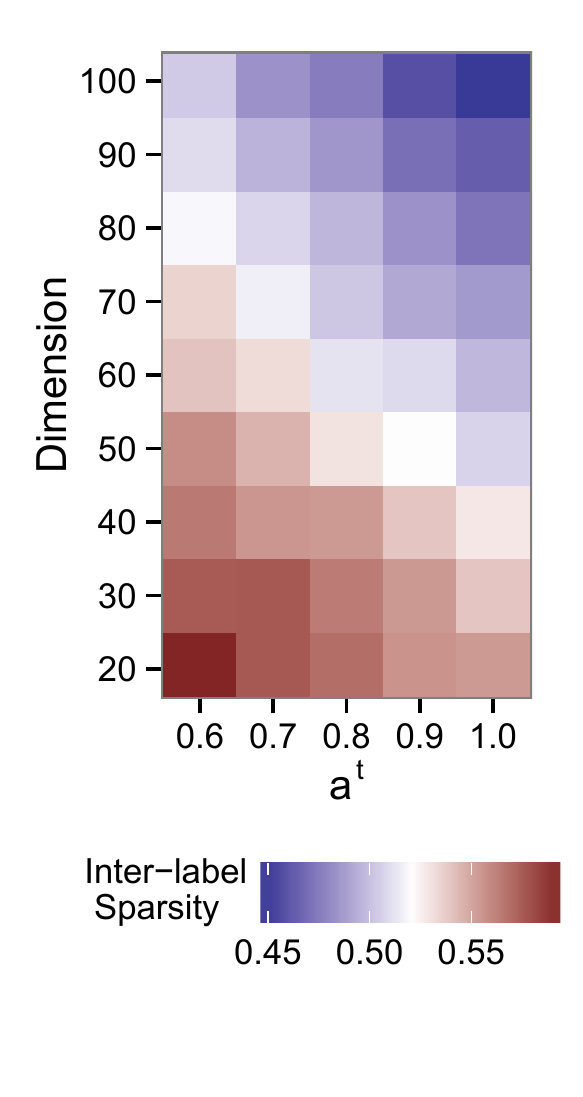}
        }%
	\end{subfigure}
         \begin{subfigure}[]{%
           \label{fig:second}
           \includegraphics[height=3.1in,valign=t]{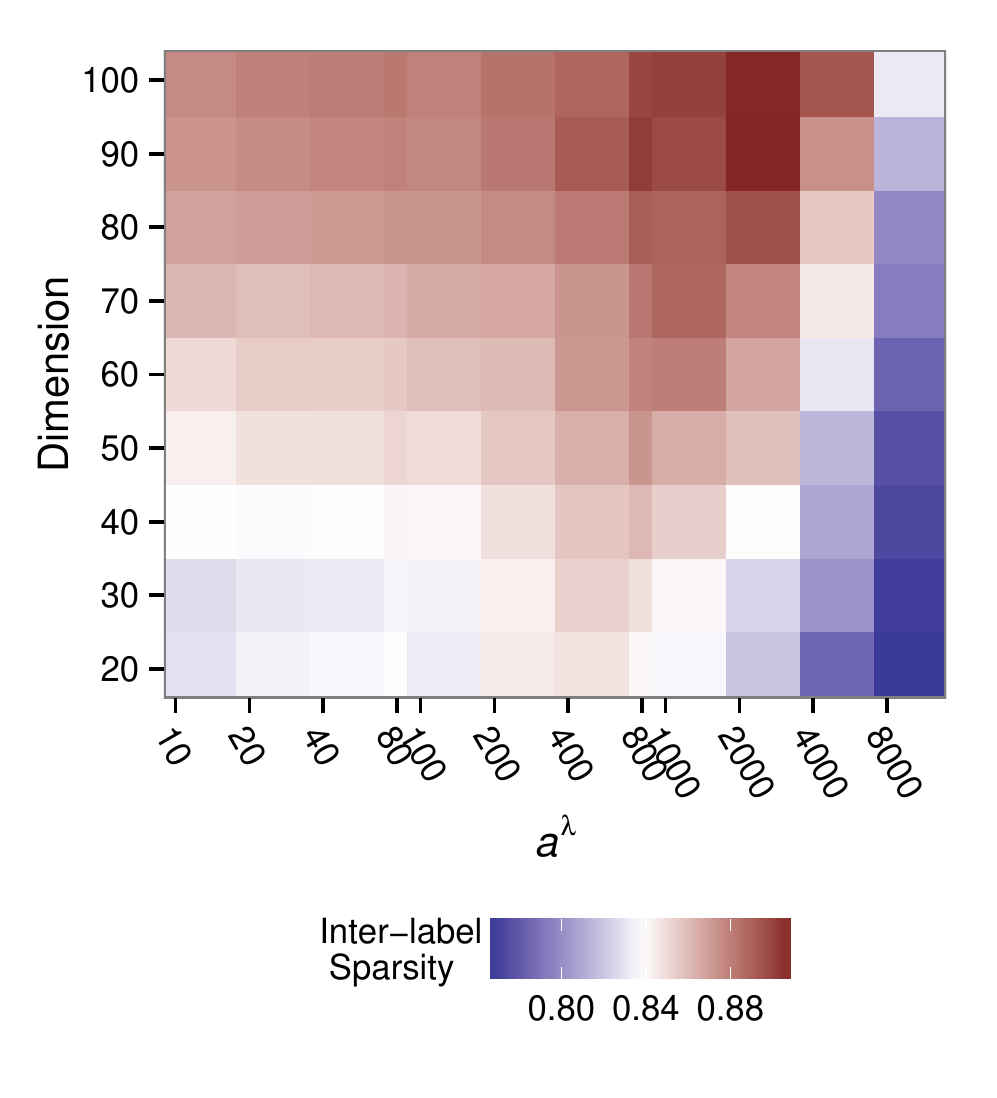}
        }
	\end{subfigure}
    \end{center}
    \caption{%
Dependence of inter-label sparsity (averaged over 10 random initializations) on dimension of representation space and parameters which control sparsity. a) Unsupervised vbNMF with sparsity constraints. b) Supervised vbNMF; inter-label sparsity can be controlled by $a^{\lambda}$.    }%
   \label{fig:subfigures}
\end{figure}
\\
To conclude the remarks on the experiments, it is worth mentioning that, if sparse representations are pursued, care is advised when choosing and optimizing hyperparameters by non-Bayesian minimization of bound. Because sparse constraints both on matrices $\boldsymbol{T}$ and $\boldsymbol{V}$ act as two competing penalizations, useful decompositions are obtained more easily by constraining only one of the matrices to be sparse - either the matrix of latent components to obtain a parts-based representation, or the matrix of coefficients to obtain a sparse representation of data. So, when optimizing the shape parameter of one of the matrices in such a manner next to a fixed hyperparameter of the other matrix which is to be made sparse, due to the automated (and, equally importantly, non-Bayesian) nature of the optimization the former may also come to describe a sparse distribution and in effect impede the desired bias toward the desired type of sparsity.
\begin{table}
\centering
\caption{Summary of experimental results}
\label{table_1}
\begin{tabular}{  c |  l | c | c | c  p{5cm} |}
\cline{2-5}
   &{\multirow{2}{*}{Algorithm}}&{\multirow{2}{*}{PCA}}&Unsupervised&Supervised\\
&&&NMF&NMF\\
\hline\hline
  Micro- & [Min,Max] &&[0.5190,0.5796]&[0.6890,\textbf{0.7418}]\\ 
  averaged & Mean &0.6330&0.5532&\textbf{0.7141}\\
Accuracy & Dimension&200&40&100 \\
  \hline
  Macro- & [Min,Max] &&[0.5033,0.5655]&[0.6758,\textbf{0.7277}]\\ 
  averaged & Mean &0.6179&0.5393&\textbf{0.6997}\\
 Accuracy & Dimension&200&40&100\\
\hline
\end{tabular}
\end{table}
\section{Conclusion}
It has been well documented that using label information in low-rank representation learning is vital to obtain representations with good discriminative properties. In this context, applied to classification of a document corpus, a probabilistic learning algorithm which combines sparse coding and supervised learning has been presented.\\
To characterize advantages of using label information, two extreme cases have been juxtaposed, the presented supervised model and a fully unsupervised one, belonging to the same family, having the same noise model and using the same metaalgorithm for parameter learning.\\
A qualitative inspection motivated the introduction of the notion of inter-label sparsity, abstracting sparsity of coefficients on the level of documents to sparsity on the level of document labels. Experiments point to a strong connection between the inter-label sparsity of the representation and the classification performance metrics. Furthermore, inter-label sparsity of decompositions obtained by supervised vbNMF can elegantly be controlled by a single parameter. However, even though sparsity and nonnegativity constraints intuitively seem appropriate and result in compact and interpretable document representations, a question remains whether there is any actual advantage in using sparse representations over dense ones as classification precursors.\\
As quality of representation spaces has been primarily addressed in this work, little regard has been given to quality of the classifier \textit{per se}. Because it is reasonable to expect that a stronger classifier would result in even better classification results, it would be interesting to compare a well-tuned classifier in the representation spaces obtained by supervised vbNMF to state-of-the-art aproaches in the field, on benchmark datasets. Future work based on semi-supervised modifications of the model is considered, to make the model more flexible and 
applicable in more commonly occuring, semi-supervised, scenarios.\\
\subsubsection{Acknowledgments}
This work was supported by the Croatian Ministry of Science, Education and Sports through the project "Computational Intelligence Methods in Measurement Systems", No. 098-0982560-2565.
\bibliographystyle{alpha}
\bibliography{MyCollection}
\newpage
\appendix
\setcounter{table}{0}
\renewcommand{\thetable}{A\arabic{table}}
\section{Probability Density Functions}
By $\mydigamma\left(.\right)$ digamma function, $\mydigamma\left(x\right)=\frac{d}{dx}\ln \Gamma (x)$, is denoted.
\subsection{Poisson Probability Density Function}
Definition:
\begin{equation}
	\mathcal{P}(x|\lambda )=e^{(-\lambda +x\ln \lambda -\ln \mydigamma(x+1))},\lambda>0\nonumber.
\end{equation}
Sufficient statistics:
\begin{equation}
	\left\langle x\right\rangle =\lambda \nonumber.
\end{equation}
\subsection{Gamma Probability Density Function}
Definition:
\begin{equation}
	\mathcal{G}\left(x|a,b\right)=e^{\left(-\frac{1}{b}x+(a-1)\ln x-a\ln b-\ln \mydigamma (a)\right)},a>0,b>0\nonumber.
\end{equation}
Sufficient statistics:
\begin{equation}
	\langle x\rangle =ab\nonumber,
\end{equation}
\begin{equation}
	\langle \ln  x\rangle =\mydigamma(a)+\ln b\nonumber.
\end{equation}
Entropy:
\begin{equation}
	\mathcal{H}\left[\mathcal{G}\left(x|a,b\right)\right]=-(a-1)\mydigamma (a)+\ln b+a+\ln \Gamma(a)\nonumber.
\end{equation}
\subsection{Multinomial Probability Density Function}
Definition:
\begin{IEEEeqnarray}{rCl}
	\IEEEeqnarraymulticol{3}{l}{
		\mathcal{M}\left(\overset{\rightharpoonup }{x} \left|s,\overset{\rightharpoonup }{p}\right.\right)
=\delta \left(s-\sum _i x_i\right)e^{\left(\ln \Gamma (s+1)+\sum _i \left(x_i\ln p_i-\ln \Gamma	\left(x_i+1\right)\right)\right)}},\nonumber\\
	&&\sum _i p_i=1,\sum _i x_i=s\nonumber.
\end{IEEEeqnarray}
Sufficient statistics:
\begin{equation}
	\left[
	\begin{array}{c}
		 \left\langle \ln x_1\right\rangle  \\
		 \vdots  \\
	 \left\langle \ln x_C\right\rangle 
	\end{array}
	\right]=s\left[
	\begin{array}{c}
		 p_1 \\
		 \vdots  \\
		 p_C
	\end{array}
	\right]\nonumber.
\end{equation}
Entropy:
\begin{IEEEeqnarray}{rCl}
	\mathcal{H}\left[\mathcal{M}\left(\overset{\rightharpoonup }{x} \left|s,\overset{\rightharpoonup }{p}\right.\right) \right]&=&-\ln \Gamma (s+1)-\sum _i \left\langle x_i\right\rangle \ln p_i\nonumber\\
	&&+\sum _i \left\langle \ln \Gamma \left(x_i+1\right)\right\rangle-\left\langle \ln \delta \left(s-\sum _i x_i\right)\right\rangle \nonumber.
\end{IEEEeqnarray}
\section{Variational Bayesian Learning Algorithm}
Iterative alternating update rules follow from (11) by plugging in
\begin{equation}
	p\left(\boldsymbol{D},\boldsymbol{\Theta}\right)=p\left(\boldsymbol{X},\boldsymbol{S},\boldsymbol{T},\boldsymbol{V},\boldsymbol{\Lambda}\left|\boldsymbol{A}^{\boldsymbol{t}}\right.,\boldsymbol{B}^{\boldsymbol{t}},\boldsymbol{A}^{\boldsymbol{v}},\boldsymbol{B}^{\boldsymbol{v}}\right)\nonumber
\end{equation}
and
\begin{equation}
	q\left(\boldsymbol{\Theta}\right)=q\left(\boldsymbol{S},\boldsymbol{T},\boldsymbol{V},\boldsymbol{\Lambda}\right).\nonumber
\end{equation}
Derivation of the iterative alternating update rules will be sketched as follows: for each instrumental distribution first its analytical form will be specified as follows from (11), followed by its natural parameters in the top and its sufficient statistics in the bottom row of the following table. Notice that the algorithm stores these distributions from iteration to iteration in form of sufficient statistics.
%
%
\begin{flalign*}
	q\left(s_{\nu :\tau }\right){}^{(t+1)}&\propto e^{\left\langle \ln p\left(\boldsymbol{X},\boldsymbol{S},\boldsymbol{T},\boldsymbol{V},		\boldsymbol{\Lambda}\middle|\boldsymbol{A^t},\boldsymbol{B^t},\boldsymbol{A^\lambda},\boldsymbol{B^\lambda},\overset{\rightharpoonup }{z}\right)\right\rangle {}_{\frac{q\left(\boldsymbol{S},\boldsymbol{T},\boldsymbol{V},\boldsymbol{\Lambda}\right)^{(t)}}{q\left(s_{\nu :\tau }\right){}^{(t)}}}}\\
&=\mathcal{M}\left(s_{\nu :\tau }|x_{\nu \tau },p_{\nu :\tau }{}^{(t)}\right),&\nonumber
\end{flalign*}
\begin{equation}
\begin{IEEEeqnarraybox}[\IEEEeqnarraystrutmode \IEEEeqnarraystrutsizeadd{12pt}{12pt}] {v/l/v}
\IEEEeqnarrayrulerow\\
&\quad p_{\nu i\tau }{}^{(t)}=\frac{\exp\left(\left\langle \ln t_{\nu i}\right\rangle {}^{(t)}+\left\langle \ln v_{i\tau }\right\rangle {}^{(t)}\right)}{\sum _i \exp\left(\left\langle \ln t_{\nu i}\right\rangle {}^{(t)}+\left\langle \ln v_{i\tau }\right\rangle {}^{(t)}\right)} \quad \\
\IEEEeqnarrayrulerow\\
&\quad \left\langle s_{\nu i\tau }\right\rangle {}^{(t+1)}=x_{\nu \tau }p_{\nu i\tau }{}^{(t)} \quad \\
\IEEEeqnarrayrulerow
\end{IEEEeqnarraybox}\nonumber
\end{equation}
\\
%
%
\begin{flalign*}
	q\left(t_{\nu i}\right){}^{(t+1)}&\propto e^{\left\langle \ln p\left(\boldsymbol{X},\boldsymbol{S},\boldsymbol{T},\boldsymbol{V},		\boldsymbol{\Lambda}\middle|\boldsymbol{A^t},\boldsymbol{B^t},\boldsymbol{A^\lambda},\boldsymbol{B^\lambda},\overset{\rightharpoonup }{z}\right)\right\rangle {}_\frac{q\left(\boldsymbol{S},\boldsymbol{T},\boldsymbol{V},\boldsymbol{\Lambda}\right)^{(t)}}{q\left(t_{\nu i}\right){}^{(t)}}}\\
&=\mathcal{G}\left(t_{\nu i}\left|\alpha _{\nu i}^t{}^{(t)},\beta _{\nu i}^t{}^{(t)}\right.\right)&
\end{flalign*}
\begin{equation}
\begin{IEEEeqnarraybox}[\IEEEeqnarraystrutmode \IEEEeqnarraystrutsizeadd{12pt}{12pt}] {v/l/v/}
\IEEEeqnarrayrulerow\\
&\quad  \alpha _{\nu i}^t{}^{(t)}=a_{\nu i}^t+\sum _{\tau } \left\langle s_{\nu i\tau }\right\rangle ^{(t)} \quad \\
&\quad  \beta _{\nu i}^t{}^{(t)}=\left(b_{\nu i}^t{}^{-1}+\sum _{\tau } \left\langle v_{i\tau }\right\rangle {}^{(t)}\right)^{-1} \quad \\
\IEEEeqnarrayrulerow\\
&\quad \left\langle t_{\nu i}\right\rangle {}^{(t+1)}=\alpha _{\nu i}^t{}^{(t)}\beta _{\nu i}^t{}^{(t)} \quad \\
&\quad  \left\langle \ln t_{\nu i}\right\rangle {}^{(t+1)}=\mydigamma \left(\alpha _{\nu i}^t{}^{(t)}\right)+\ln \beta _{\nu i}^t{}^{(t)} \quad \\
\IEEEeqnarrayrulerow
\end{IEEEeqnarraybox}\nonumber
\end{equation}
%
%
\begin{flalign*}
	q\left(v_{i\tau}\right){}^{(t+1)}&\propto e^{\left\langle \ln p\left(\boldsymbol{X},\boldsymbol{S},\boldsymbol{T},\boldsymbol{V}\left|\boldsymbol{A}^{\boldsymbol{t}}\right.,\boldsymbol{B}^{\boldsymbol{t}},
\boldsymbol{A}^{\boldsymbol{v}},\boldsymbol{B}^{\boldsymbol{v}}\right)\right\rangle {}_\frac{q\left(\boldsymbol{S},\boldsymbol{T},\boldsymbol{V},\boldsymbol{\Lambda}\right)^{(t)}}{q\left(v_{i\tau}\right){}^{(t)}}}\\
&=\mathcal{G}\left(v_{i\tau}\left|\alpha _{i\tau}^v{}^{(t)},\beta _{i\tau}^v{}^{(t)}\right.\right)&
\end{flalign*}
\begin{equation}
\begin{IEEEeqnarraybox}[\IEEEeqnarraystrutmode \IEEEeqnarraystrutsizeadd{12pt}{12pt}] {v/l/v/}
\IEEEeqnarrayrulerow\\
&\quad \alpha _{i\tau}^v{}^{(t)}=1+\sum _{\nu} \left\langle s_{\nu i\tau }\right\rangle ^{(t)}\quad  \\
&\quad \beta _{i\tau}^v{}^{(t)}=\left(\sum_{l} \dirac(z_\tau-l) \left\langle\lambda_{i l} \right\rangle^{(t)}+\sum _{\nu } \left\langle t_{\nu i }\right\rangle {}^{(t)}\right)^{-1} \quad\\
\IEEEeqnarrayrulerow\\
&\quad \left\langle v_{i\tau}\right\rangle {}^{(t+1)}=\alpha _{i\tau}^v{}^{(t)}\beta _{i\tau}^v{}^{(t)} \quad  \\
&\quad \left\langle \ln v_{i\tau}\right\rangle {}^{(t+1)}=\mydigamma \left(\alpha _{i\tau}^v{}^{(t)}\right)+\ln \beta _{i\tau}^v{}^{(t)} \quad \\
\IEEEeqnarrayrulerow
\end{IEEEeqnarraybox}\nonumber
\end{equation}
\\ \\
Update expressions for $q\left(\lambda_{il}\right){}^{(t+1)}$ depend on a difficult term in the bound. Based on concavity of logaritmic function, this term can be lower bounded using Jensen's inequality:
\begin{equation}
\left\langle \ln \left(\sum_{l} \dirac(z_\tau-l) \lambda_{i l}\right) \right\rangle{}_\frac{q\left(\boldsymbol{S},\boldsymbol{T},\boldsymbol{V},\boldsymbol{\Lambda}\right)^{(t)}}{q\left(\lambda_{il}\right){}^{(t)}}\geq \sum_{l}\dirac(z_\tau-l)\left\langle \ln  \lambda_{i l}\right\rangle{}_\frac{q\left(\boldsymbol{S},\boldsymbol{T},\boldsymbol{V},\boldsymbol{\Lambda}\right)^{(t)}}{q\left(\lambda_{il}\right){}^{(t)}}.\nonumber
\end{equation}
In this relaxed bound convergence is preserved and now updates have analytical forms:
\begin{flalign*}
	q\left(\lambda_{il}\right){}^{(t+1)} &\propto e^{\left\langle \ln p\left(\boldsymbol{X},\boldsymbol{S},\boldsymbol{T},\boldsymbol{V},	\boldsymbol{\Lambda}\middle|\boldsymbol{A^t},\boldsymbol{B^t},\boldsymbol{A^\lambda},\boldsymbol{B^\lambda},\overset{\rightharpoonup }{z}\right)\right\rangle {}_\frac{q\left(\boldsymbol{S},\boldsymbol{T},\boldsymbol{V},\boldsymbol{\Lambda}\right)^{(t)}}{q\left(\lambda_{il}\right){}^{(t)}}}\\
&=\mathcal{G}\left(v_{il}\left|\alpha _{il}^\lambda{}^{(t)},\beta _{il}^\lambda{}^{(t)}\right.\right)&
\end{flalign*}
\begin{equation}
\begin{IEEEeqnarraybox}[\IEEEeqnarraystrutmode \IEEEeqnarraystrutsizeadd{12pt}{12pt}] {v/l/v/}
\IEEEeqnarrayrulerow\\
&\quad \alpha _{il}^\lambda{}^{(t)}=a_{il}^\lambda{}+\sum_{l} \dirac(z_\tau-l) \quad   \\
&\quad \beta _{il}^\lambda{}^{(t)}=\left( b_{il}^\lambda{}^{-1}+\sum_{\tau} \left\langle  v_{i\tau}\right\rangle {}^{(t)} \dirac(z_\tau-l)\right)^{-1} \quad\\
\IEEEeqnarrayrulerow\\
&\quad \left\langle \lambda_{il}\right\rangle {}^{(t+1)}=\alpha _{il}^\lambda{}^{(t)}\beta _{il}^\lambda{}^{(t)} \quad  \\
&\quad \left\langle \ln \lambda_{il}\right\rangle {}^{(t+1)}=\mydigamma \left(\alpha _{il}^\lambda{}^{(t)}\right)+\ln \beta _{il}^\lambda{}^{(t)} \quad \\
\IEEEeqnarrayrulerow
\end{IEEEeqnarraybox}\nonumber
\end{equation}
\\
%
The algorithm can be rewritten in matrix form: using matrix notation from Table A1. and Table A2., the learning algorithm is summarized in Table A3., where by $.*$ and $./$ elementwise matrix product and elementwise matrix division are denoted, respectively, and by $\boldsymbol{1}$ matrix of ones of appropriate dimensions.\\
Lower bound can be derived from (10) and used as convergence indicator and also as a basis for model comparison.
\begin{table}
\centering
\caption{Observed variables and hyperparameters of supervised vbNMF model}
\label{table_1}
\begin{IEEEeqnarraybox}[\IEEEeqnarraystrutmode \IEEEeqnarraystrutsizeadd{8pt}{5pt}] {v/l/v/l/v/l/v}
\IEEEeqnarrayrulerow\\
&\left[\boldsymbol{X}\right]_{\nu \tau }=x_{\nu \tau }&&\left[\boldsymbol{A}^{\boldsymbol{t}}\right]_{\nu i}=a_{\nu i}^t&&\left[\boldsymbol{A}^{\boldsymbol{\lambda}}\right]_{il }=a_{il}^\lambda\\
&\left[\boldsymbol{\Delta}\right]_{\tau l}=\dirac\left(z_\tau-l\right)&&\left[\boldsymbol{B}^{\boldsymbol{t}}\right]_{\nu i}=b_{\nu i}^t&&\left[\boldsymbol{B}^{\boldsymbol{\lambda}}\right]_{il }=b_{il}^\lambda\hfill&\\
\IEEEeqnarrayrulerow
\end{IEEEeqnarraybox}
\end{table}\begin{table}
\centering
\caption{Variational parameters of supervised vbNMF model}
\label{table_2}
\begin{IEEEeqnarraybox}[\IEEEeqnarraystrutmode \IEEEeqnarraystrutsizeadd{8pt}{5pt}] {v/l/v/l/v}
\IEEEeqnarrayrulerow\\
&\left[\boldsymbol{E}_{\boldsymbol{t}}^{(t)}\right]_{\nu i}=\left\langle t_{\nu i}\right\rangle ^{(t)}&&\left[\boldsymbol{E}_{\boldsymbol{v}}^{(t)}\right]_{i\tau }=\left\langle v_{i\tau }\right\rangle^{(t)} \\
&\left[\boldsymbol{L}_{\boldsymbol{t}}^{(t)}\right]{}_{\nu i}=\left\langle \ln t_{\nu i}\right\rangle {}^{(t)}&&\left[\boldsymbol{L}_{\boldsymbol{v}}^{(t)}\right]{}_{i\tau }=\left\langle \ln v_{i\tau }\right\rangle {}^{(t)}&\\
\IEEEeqnarrayrulerow\\
&\left[\boldsymbol{\Sigma}_{\boldsymbol{t}}^{(t)}\right]_{\nu i}=\sum _{\tau } \left\langle s_{\nu i\tau }\right\rangle^{(t)}&&\left[\boldsymbol{E}_{\boldsymbol{\lambda}}^{(t)}\right]_{i l}=\left\langle \lambda_{il}\right\rangle ^{(t)}\\ 
&\left[\boldsymbol{\Sigma}_{\boldsymbol{v}}^{(t)}\right]_{i\tau }=\sum _{\nu } \left\langle s_{\nu i\tau }\right\rangle^{(t)}&&\left[\boldsymbol{L}_{\boldsymbol{\lambda}}^{(t)}\right]_{i l}=\left\langle \ln \lambda_{i l}\right\rangle ^{(t)}\\
\IEEEeqnarrayrulerow
\end{IEEEeqnarraybox}
\end{table}\begin{table}
\centering
\caption{The learning algorithm in matrix form}
\label{table_3}
\begin{IEEEeqnarraybox}[\IEEEeqnarraystrutmode \IEEEeqnarraystrutsizeadd{5pt}{5pt}]{v/l/l/v}
\IEEEeqnarrayrulerow\\
&	\text{Inputs:} &\\
&&\boldsymbol{X},\boldsymbol{A^t},\boldsymbol{B^t},\boldsymbol{A^\lambda},\boldsymbol{B^\lambda},\boldsymbol{\Delta}\\
&	\text{Initialize:} &\\
&& \boldsymbol{E}_{\boldsymbol{t}}^{(0)},\boldsymbol{L}_{\boldsymbol{t}}^{(0)},
\boldsymbol{E}_{\boldsymbol{v}}^{(0)},\boldsymbol{L}_{\boldsymbol{v}}^{(0)},
\boldsymbol{\Sigma}_{\boldsymbol{t}}^{(0)},\boldsymbol{\Sigma}_{\boldsymbol{v}}^{(0)},
\boldsymbol{E}_{\boldsymbol{\lambda}}^{(0)},\boldsymbol{L}_{\boldsymbol{\lambda}}^{(0)}\\
&& t=0\\
&	\text{Loop:}&\\
&&	\boldsymbol{\Xi}=\boldsymbol{X}.\left/\left(\left(\exp\boldsymbol{L}_{\boldsymbol{t}}^{(t)}\right)*\left(\exp\boldsymbol{L}_{\boldsymbol{v}}^{(t)}\right)\right)\right.\nonumber
\\
&&\boldsymbol{\Sigma}_{\boldsymbol{v}}^{(t+1)}=\exp \boldsymbol{L}_{\boldsymbol{v}}^{(t)}.*\left( \left(\exp\boldsymbol{L}_{\boldsymbol{t}}^{(t)}\right)^T*\boldsymbol{\Xi} \right)\nonumber
\\
&&	\boldsymbol{\Sigma}_{\boldsymbol{t}}^{(t+1)}=\exp \boldsymbol{L}_{\boldsymbol{t}}^{(t)}.*\left(\boldsymbol{\Xi}*\left(\exp\boldsymbol{L}_{\boldsymbol{v}}^{(t)}\right)^T\right)\nonumber
\\
&&\boldsymbol{A}_{\boldsymbol{t}}=\boldsymbol{A}^{\boldsymbol{t}}+\boldsymbol{\Sigma}_{\boldsymbol{t}}^{(t+1)}\nonumber
\\
&&	\boldsymbol{B}_{\boldsymbol{t}}=1.\left/\left(1.\left/\boldsymbol{B}^{\boldsymbol{t}}\right.+\boldsymbol{1}*\left(\boldsymbol{E}_{\boldsymbol{v}}^{(t)}\right)^T\right)\right.\nonumber
\\
&&	\boldsymbol{E}_{\boldsymbol{t}}^{(t+1)}=\boldsymbol{A}_{\boldsymbol{t}}.*\boldsymbol{B}_{\boldsymbol{t}}\nonumber
\\
&&	\boldsymbol{L}_{\boldsymbol{t}}^{(t+1)}=\mydigamma \left(\boldsymbol{A}_{\boldsymbol{t}}\right)+\ln \boldsymbol{B}_{\boldsymbol{t}}\nonumber\\
&&\boldsymbol{A}_{\boldsymbol{v}}=\boldsymbol{1}+\boldsymbol{\Sigma}_{\boldsymbol{v}}^{(t+1)}\nonumber
\\
&&	\boldsymbol{B}_{\boldsymbol{v}}=1.\left/\left(\boldsymbol{E}_{\boldsymbol{\lambda}}^{(t)}*\boldsymbol{\Delta}+\left(\boldsymbol{E}_{\boldsymbol{t}}^{(t)}\right)^T*\boldsymbol{1}\right)\right.\nonumber\\
&&	\boldsymbol{E}_{\boldsymbol{v}}^{(t+1)}=\boldsymbol{A}_{\boldsymbol{v}}.*\boldsymbol{B}_{\boldsymbol{v}}\nonumber\\
&&	\boldsymbol{L}_{\boldsymbol{v}}^{(t+1)}=\mydigamma \left(\boldsymbol{A}_{\boldsymbol{v}}\right)+\ln \boldsymbol{B}_{\boldsymbol{v}}\nonumber\\
&&\boldsymbol{A}_{\boldsymbol{\lambda}}=\boldsymbol{A}^{\boldsymbol{\lambda}}+\boldsymbol{\Delta}*\boldsymbol{1}\nonumber
\\
&&	\boldsymbol{B}_{\boldsymbol{\lambda}}=1.\left/\left(1.\left/\boldsymbol{B}^{\boldsymbol{\lambda}}\right.+\boldsymbol{E}_{\boldsymbol{v}}^{(t+1)}.*\boldsymbol{\Delta}\right)\right.\nonumber\\
&&	\boldsymbol{E}_{\boldsymbol{\lambda}}^{(t+1)}=\boldsymbol{A}_{\boldsymbol{\lambda}}.*\boldsymbol{B}_{\boldsymbol{\lambda}}\nonumber\\
&&	\boldsymbol{L}_{\boldsymbol{\lambda}}^{(t+1)}=\mydigamma \left(\boldsymbol{A}_{\boldsymbol{\lambda}}\right)+\ln \boldsymbol{B}_{\boldsymbol{\lambda}}\nonumber\\
&&	\text{Hyperparameter optimization (non-Bayesian)}\\
&	\text{End loop}& \\
\IEEEeqnarrayrulerow
\end{IEEEeqnarraybox}
\end{table}
\end{document}